\documentclass[letterpaper, 10 pt, conference]{ieeeconf}  

\IEEEoverridecommandlockouts                              

\title{\LARGE \bf
Octopus-like Reaching Motion: A Perspective Inspired by Whipping
}

\author{%
Shengyao Zhang\textsuperscript{\dag}, 
Yiyuan Zhang\textsuperscript{\dag,*}, \textit{Graduate Student Member, IEEE},\\
Chenrui Zhang, \textit{Graduate Student Member, IEEE}, 
Yiming Li, 
Wenci Xin, \\
Yuliang Liufu, 
Hong Wei Ng, 
and Cecilia Laschi, \textit{Fellow, IEEE}%
\thanks{\textsuperscript{\dag}These authors contributed equally to this work.}%
\thanks{*Corresponding author: Yiyuan Zhang (yiyuan.zhang@u.nus.edu)}%
\thanks{Department of Mechanical Engineering, National University of Singapore, Singapore 117575, Singapore.}%
\thanks{This work was supported by the following projects: the DESTRO project (“Dextrous, strong yet soft robots,” R2210IR124), funded by MAE (Italy) and A*STAR (Singapore); the REBOT project (“Rethinking underwater robot manipulation,” Moe-t2eP50221-0010), funded by the Singapore Ministry of Education; and the RoboLife project (“Soft robots with morphological adaptation and life-like abilities”), funded by the NUS Funding Agency (Singapore).The authors also would like to thank Wanlin Huang for help with  final result visualization.}%
}

\usepackage{graphicx}     
\usepackage[caption=false,font=footnotesize]{subfig}
\usepackage{stfloats}
\usepackage{cite}

\begin{document}

\maketitle
\thispagestyle{empty}
\pagestyle{empty}

\begin{abstract}

The stereotypical reaching motion of the octopus arm has drawn growing attention for its efficient control of a highly deformable body. Previous studies suggest that its characteristic bend propagation may share underlying principles with the dynamics of a whip. This work investigates whether whip-like passive dynamics in water can reproduce the kinematic features observed in biological reaching and their similarities and differences. Platform-based whipping tests were performed in water and air while systematically varying material stiffness and driving speed. Image-based quantification revealed that the Ecoflex Gel 2 arm driven at 150 rpm (motor speed) reproduced curvature propagation similar to that observed in octopus reaching. However, its bend-point velocity decreased monotonically rather than exhibiting the biological bell-shaped profile, confirming that the octopus reaching movement is not merely a passive whipping behavior. The absence of propagation in air further highlights the critical role of the surrounding medium in forming octopus-like reaching motion. This study provides a new perspective for understand biological reaching movement, and offers a potential platform for future hydrodynamic research.

\end{abstract}

\section{INTRODUCTION}

The reaching movement of the octopus arm exemplifies one of the most stereotyped and simplified control strategies in its highly deformable muscular system. Unlike the human arm with a limited number of joints, the octopus arm possesses virtually infinite degrees of freedom, functioning as a muscular hydrostat capable of bending, elongating, shortening, and twisting without skeletal support. Despite this structural complexity, octopuses can extend their arms toward a target through highly similar reaching movements, following a consistent bend-propagation pattern that is organized both spatially and temporally \cite{gutfreund1996organization,sumbre2001control,yekutieli2005dynamic}.

\begin{figure}[!b]
  \centering
  \subfloat[]{%
    \label{fig:intro_a}
    \includegraphics[width=0.9\linewidth]{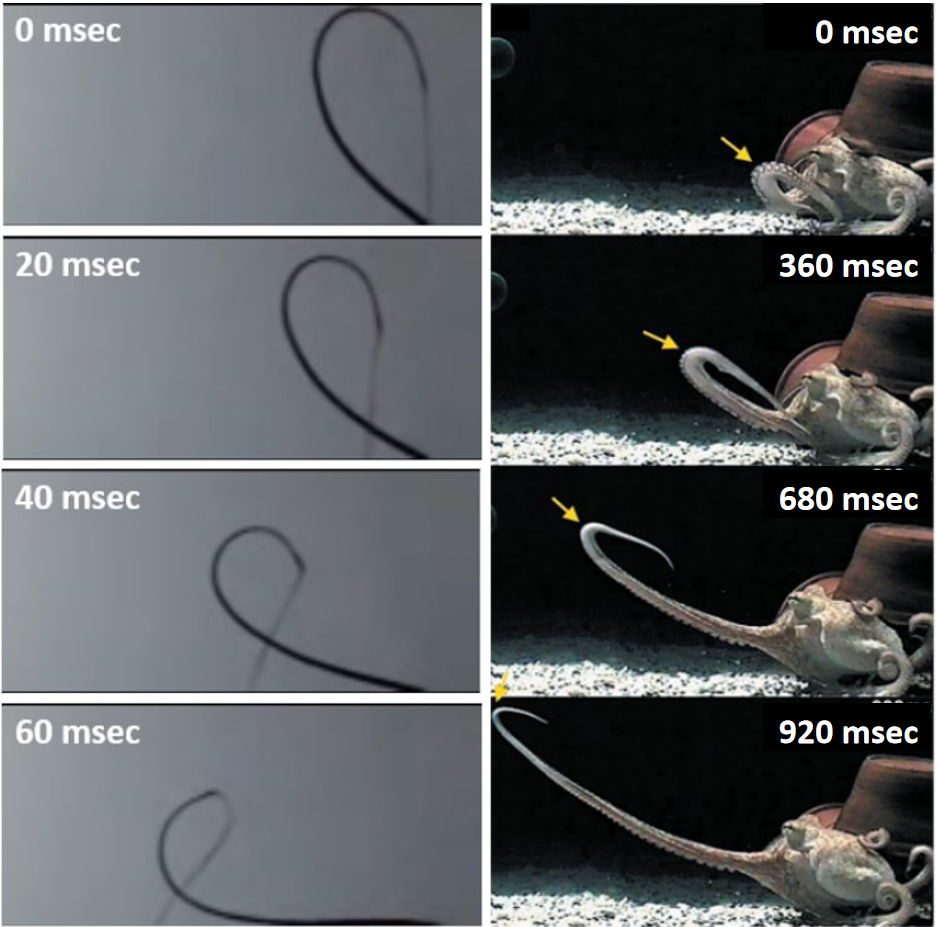}}%
 \\
  \subfloat[]{%
    \label{fig:intro_b}
    \includegraphics[width=0.9\linewidth]{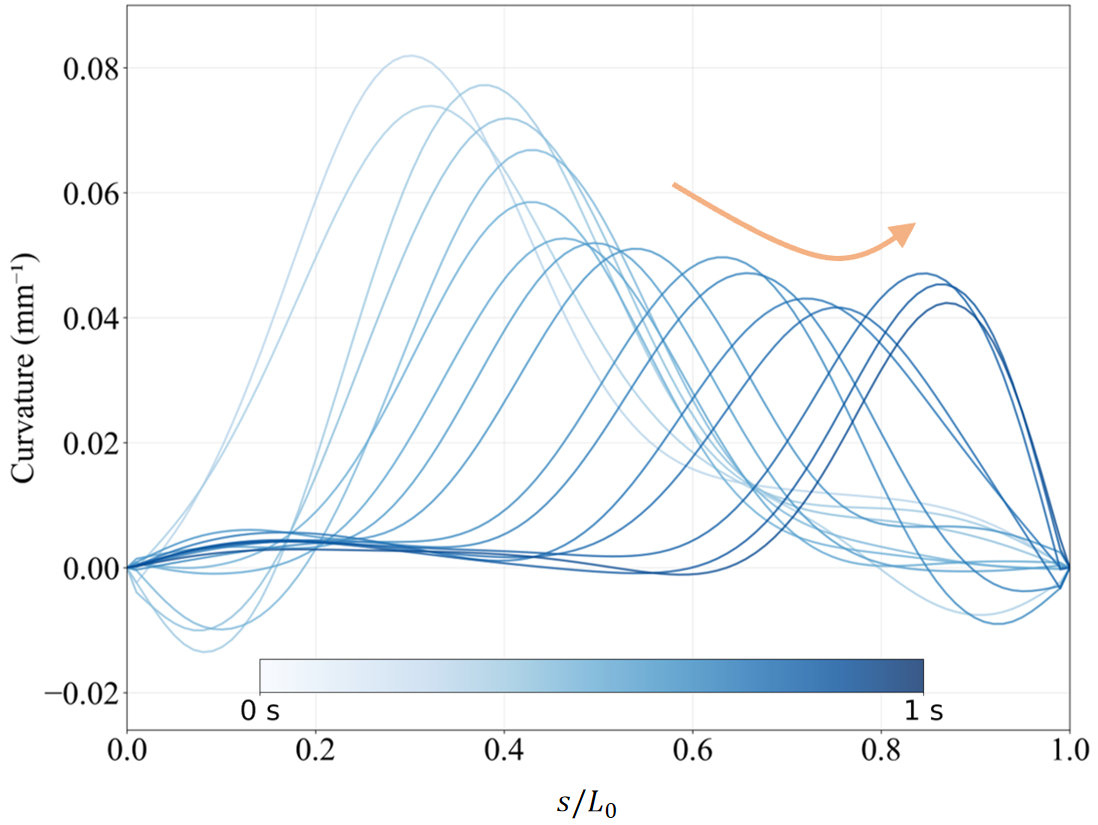}}%
  \caption{(a) Multiple exposures of a bullwhip exhibiting propagating waves during motion \cite{WorldWideWhips}, kinematically similar to an octopus arm (adapted with permission from \cite{sumbre2001control}, \textcopyright{} 2001 AAAS). Note: Image layout and comparative analysis inspired by the study in \cite{noel2018grip}. (b) Curvature pattern of real octopus reaching behavior (\(s/L_0\) denotes the normalized arc length along the arm, with \(s\) being the arc-length coordinate and \(L_0\) the undeformed arm length).}
  \label{fig:intro_pair}
\end{figure}

Previous studies have suggested that this movement shares intriguing parallels with whipping-like passive dynamics \cite{gutfreund1996organization,sumbre2001control,gutfreund1998patterns,alupay2023syntactic}. Biological reaching can be described as a base-to-tip propagating bend wave—arising from continuous neural drive together with intrinsic elasticity and hydrodynamic interaction—in which a traveling front of stiffening and muscle activation produces curvature and energy transmission \cite{zelman2013kinematic,wang2022control} reminiscent of a whip crack, as shown in Figure.~\ref{fig:intro_a} \cite{noel2018grip}. The analogy suggests that the octopus may utilize passive mechanical whipping to achieve rapid, efficient reaching movement, converting local activation into a large-scale sweeping-like reaching motion.

The detailed kinematics of the octopus reaching behavior reveal several invariant features across individuals and contexts. The motion is initiated by the formation of a localized bend near the arm base, which then propagates distally along a nearly planar trajectory, maintaining a constant-curvature bend propagation region over a significant portion of the movement, as shown in Figure.~\ref{fig:intro_b} \cite{gutfreund1996organization,yekutieli2005dynamic,zelman2013kinematic}. The bend-point velocity shows a characteristic bell-shaped temporal profile with smooth acceleration and deceleration, implying that control is exerted chiefly over the bend’s spatiotemporal position rather than over discrete muscle segments \cite{gutfreund1996organization, gutfreund1998patterns}. Biomechanical and computational models further demonstrate that a simple traveling wave of muscle stiffening can reproduce the experimentally observed velocity profiles and curvature evolutions \cite{yekutieli2005dynamic,zelman2013kinematic}, supporting the notion that the system relies on local passive wave mechanics to generate globally coordinated motion. Collectively, these findings portray octopus reaching as a dynamic balance between active control and passive mechanical propagation, embodying principles that resonate strongly with whipping dynamics—where a localized deformation evolves into a smooth, accelerating wave culminating in effective, directed motion.

The whipping motion represents a canonical example of energy focusing and curvature propagation in flexible continua. When an impulse is applied to the handle of a whip, a traveling loop or curvature wave propagates toward the free end, gradually accelerating as the whip’s radius tapers and mass per unit length decreases \cite{mcmillen2003whip}. This process converts distributed elastic and kinetic energy into a concentrated, high-velocity motion at the tip—eventually reaching supersonic speed and producing the characteristic crack. Theoretical analyses based on elastic rod models reveal that the traveling wave maintains a quasi-constant curvature while local velocity increases, consistent with conservation of energy and momentum in a homogeneous medium. Importantly, this bend propagation mechanism is not unique to whips; similar dynamic principles have been identified in biological systems such as fly-fishing (a type of fishing involving casting a lightweight lure) lines, flagella, and even octopus arms, where a local curvature travels through the body to produce directed motion. From this perspective, the octopus reaching behavior may be hypothesis as a biological manifestation of the whipping phenomenon \cite{noel2018grip}. Both systems rely on the propagation of a localized bend—maintaining nearly constant curvature and exhibiting a bell-shaped velocity evolution. In whips, the traveling loop is a fully passive mechanical wave driven by the initial impulse and geometric tapering; in octopus arms, a similar propagation arises from the coupling of muscle activation waves and passive environmental interactions. These parallels suggest that understanding whipping dynamics provides not only a physical analogue but also an experimental framework for interpreting the passive wave-based control strategy observed in octopus reaching motions.

Building upon these findings, we aimed to examine whether whipping dynamics in a fluid environment could reproduce the bend-propagation characteristics observed in octopus reaching, and whether such motions exhibit kinematic features comparable to biological reaching movements. A prevalent view is that octopus reaching is driven by active muscular control, with most existing research focusing on actively driven motion \cite{yekutieli2005dynamic,hanassy2015stereotypical, gutfreund1996organization, sumbre2001control}. Relatively few studies have investigated passively driven arm movement underwater, and none have employed controlled underwater whipping as an experimental analogy—an intuitively compelling comparison, especially to an octopus’s reaching motion. In particular, this study explores whether a whip-like passive motion, when performed underwater, can generate bend propagation, curvature evolution, and bend-point velocity profiles similar to those recorded in octopus reaching. We argue that examining such passive dynamics remains valuable for elucidating how environmental interactions, absent sustained neural activation, contribute to or constrain bend propagation—offering a complementary perspective to purely active models of movement generation.

\section{METHOD}

\subsection{Design of Mechanism}

\begin{figure*}[!t]
    \centering
    \includegraphics[width=\textwidth]{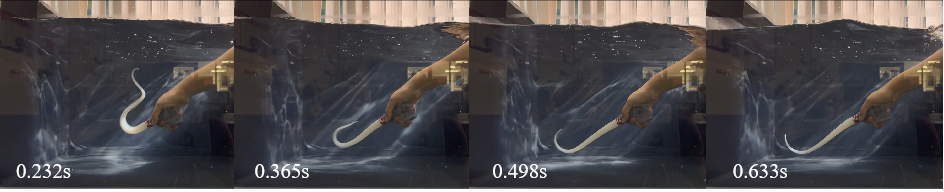}
    \caption{Underwater manual whipping motion.}
    \label{fig:Fig2}
\end{figure*}

\begin{figure}[!t]
  \centering
  \subfloat[]{%
    \label{fig:Fig3a}
    \includegraphics[width=0.48\linewidth]{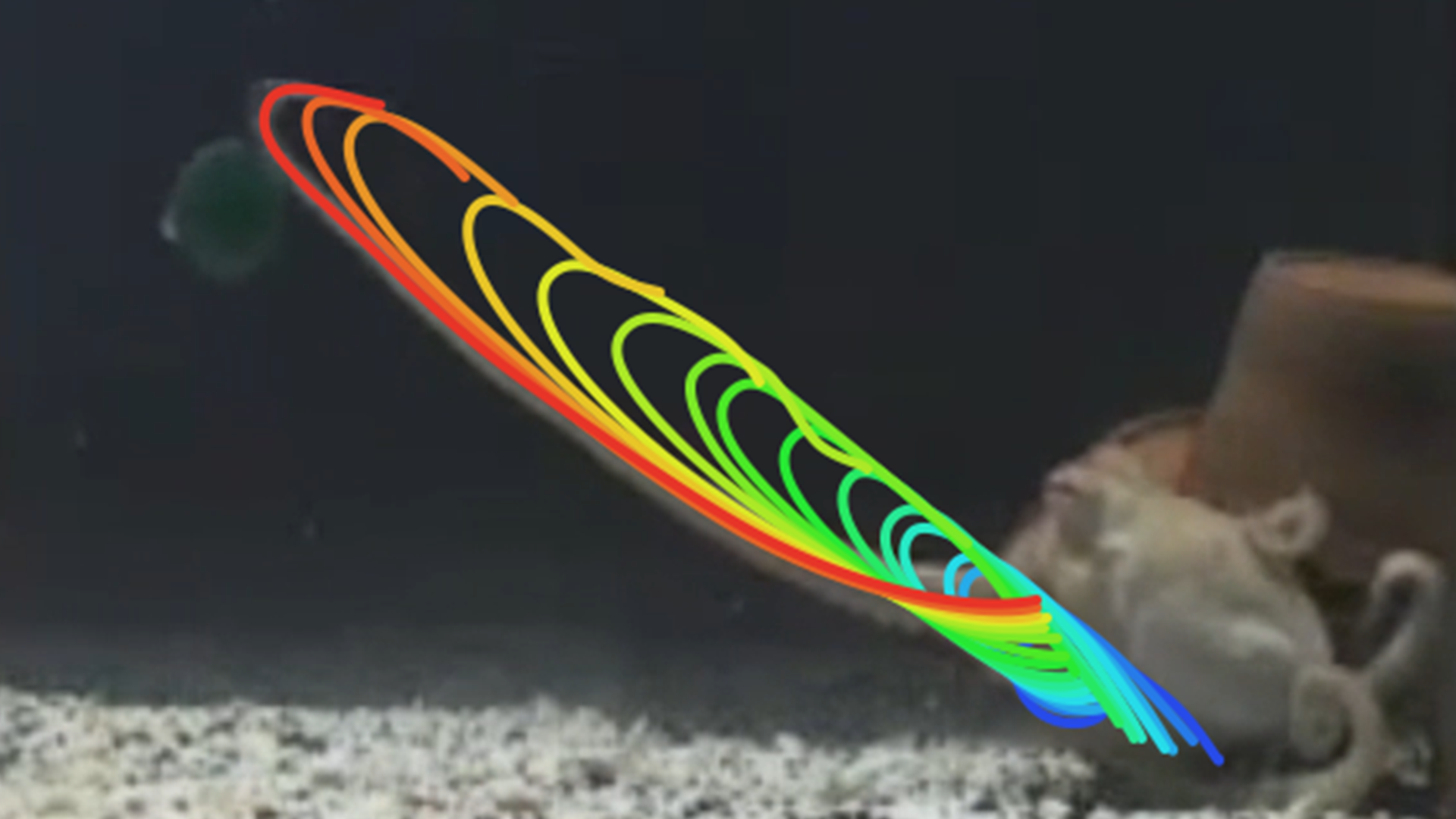}}%
  \hfill
  \subfloat[]{%
    \label{fig:Fig3b}
    \includegraphics[width=0.48\linewidth]{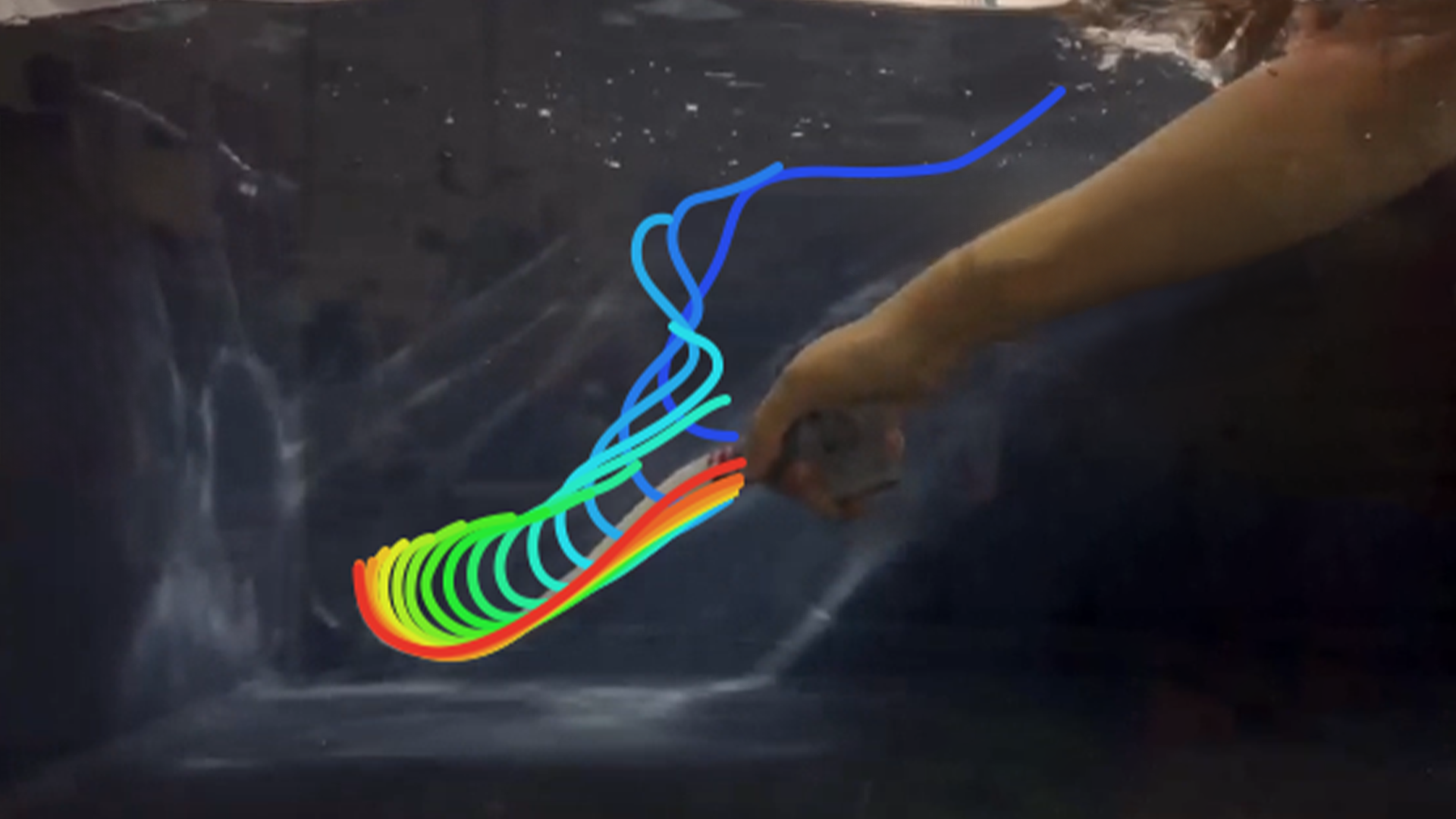}}%
  \vspace{-3pt}
  \centering
  \subfloat[]{%
    \label{fig:Fig3c}
    \includegraphics[width=0.48\linewidth]{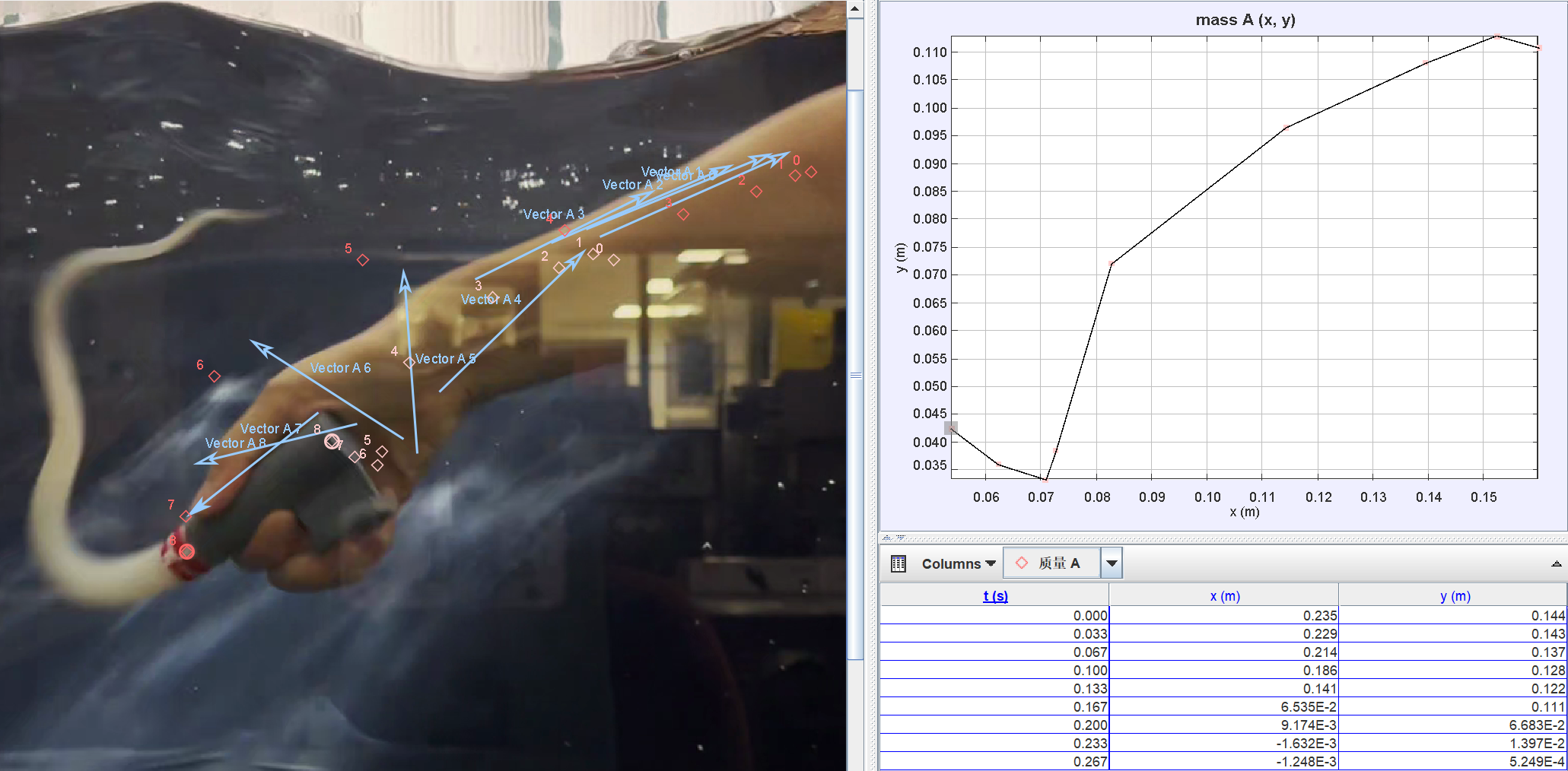}}%
  \hfill
  \subfloat[]{%
    \label{fig:Fig3d}
    \includegraphics[width=0.48\linewidth]{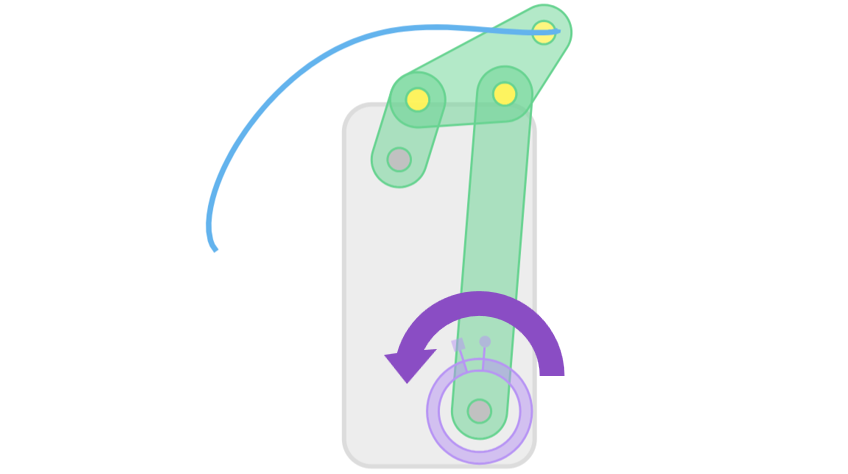}}%
  \vspace{-3pt}
  \centering
  \subfloat[]{%
    \label{fig:Fig3e}
    \includegraphics[width=0.48\linewidth]{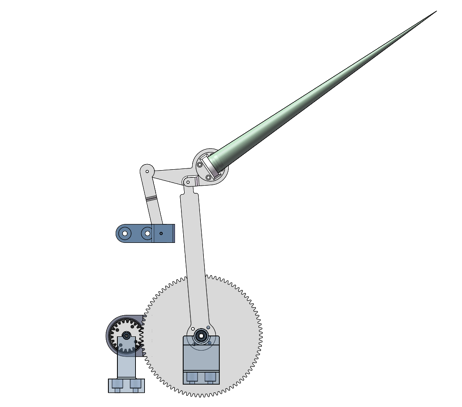}}%
  \hfill
  \subfloat[]{%
    \label{fig:Fig3f}
    \includegraphics[width=0.48\linewidth]{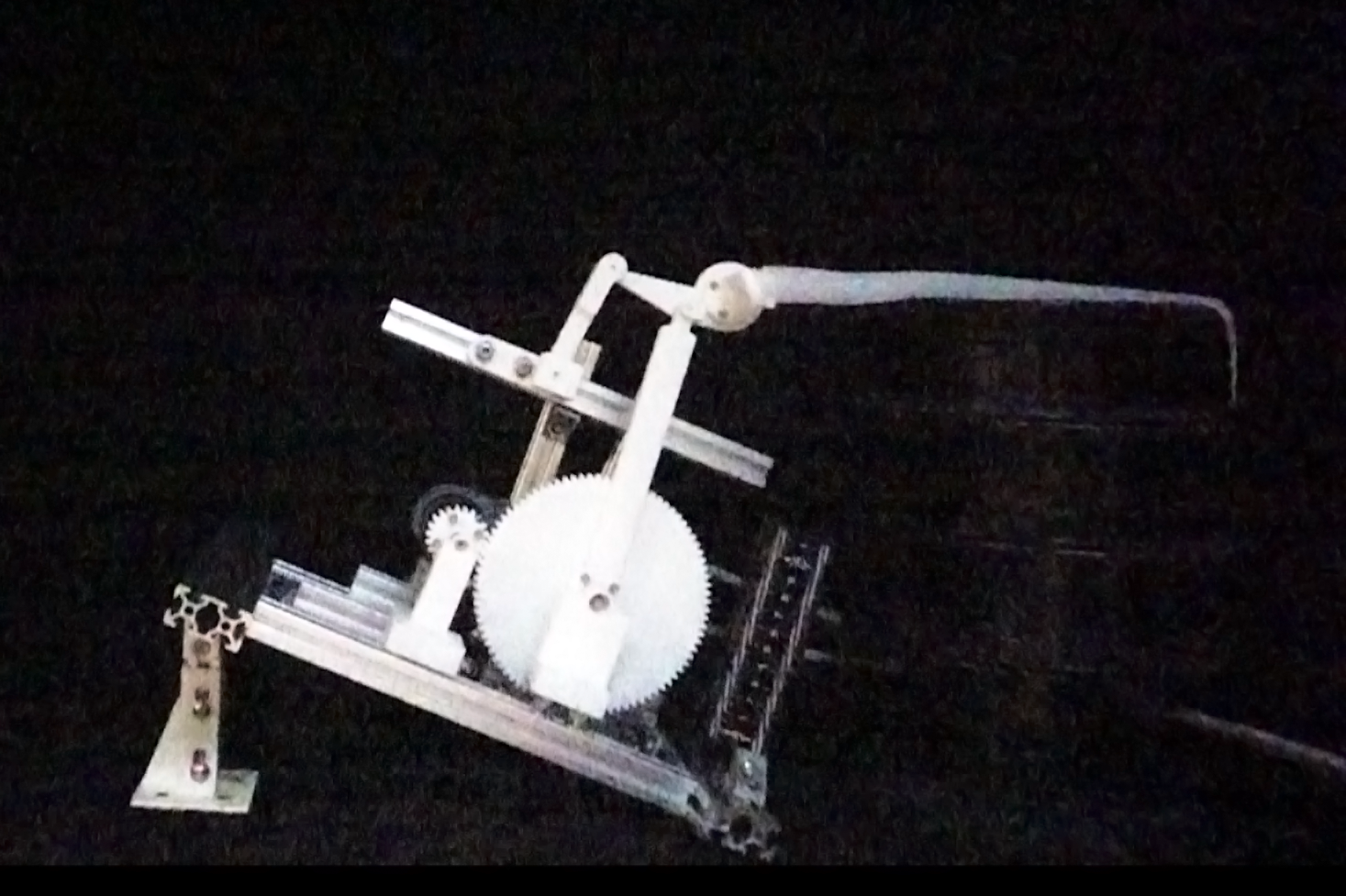}}%
 \caption{(a) The trajectory of the midline shift during the octopus's reaching motion. (b) The trajectory of the midline shift during underwater manual whipping motion. (c) Position and orientation extraction in Tracker. (d) Mechanisms generation in MotionGen. (e) SolidWorks Model. (f) Physical prototype of the mechanism.}
  \label{fig:Fig3}
\end{figure}

We first conducted manual whipping experiments in an underwater environment to qualitatively reproduce the reaching motion observed in octopus arms, as shown in Figure.~\ref{fig:Fig2}. As the extracted midline of real octopus reaching movement and manual whipping motion shown in Figure.~\ref{fig:Fig3a} and Figure.~\ref{fig:Fig3b}, the resulting motion sequence exhibited a distinct bend propagation pattern similar to the biological reaching behavior. Subsequently, a motion tracking software (Tracker) was used to extract the positional coordinates and orientation angles of the silicone arm’s base movement, as presented in Figure.~\ref{fig:Fig3c}. The recorded data were plotted on an x–y coordinate system and imported into MotionGen, where a virtual mechanism capable of outputting an analogous trajectory was generated, as shown in Figure.~\ref{fig:Fig3d}.
Based on the generated trajectory, the detailed structure of the linkage mechanism was designed in SolidWorks, including each connecting link, joint, and gear configuration (see Figure.~\ref{fig:Fig3e}). To overcome the inherent limitation of the motor’s output torque, an additional set of power-amplifying gears was incorporated. Due to the constraints of manual manipulation within the water tank, the observed motion primarily exhibited a diagonally downward wave propagation. To better approximate the actual reaching kinematics of the octopus—where the curvature wave propagates nearly along a horizontal plane—we elevated a section of the mechanism’s supporting frame by approximately 20 degrees clockwise. This adjustment redirected the path of motion transmission in a nearly horizontal direction, allowing a more biologically consistent wave propagation pattern.

Finally, all parts were fabricated using 3D printing with PLA+ material and subsequently assembled into a complete prototype mechanism with a tapered silicone arm, as shown in Figure.~\ref{fig:Fig3f} and the motor used in the mechanism has a KV value of 350 rpm/V, a maximum current of 15 A, and a maximum power of 300 W. The resulting system provided a compact platform that allows fine-tuning of geometric and actuation parameters for future experiments on wave dynamics and underwater performance.

\subsection{Data Extraction}
\begin{figure}[!b]
  \centering
  \subfloat[]{%
    \label{fig:Fig4a}
    \includegraphics[width=0.48\linewidth]{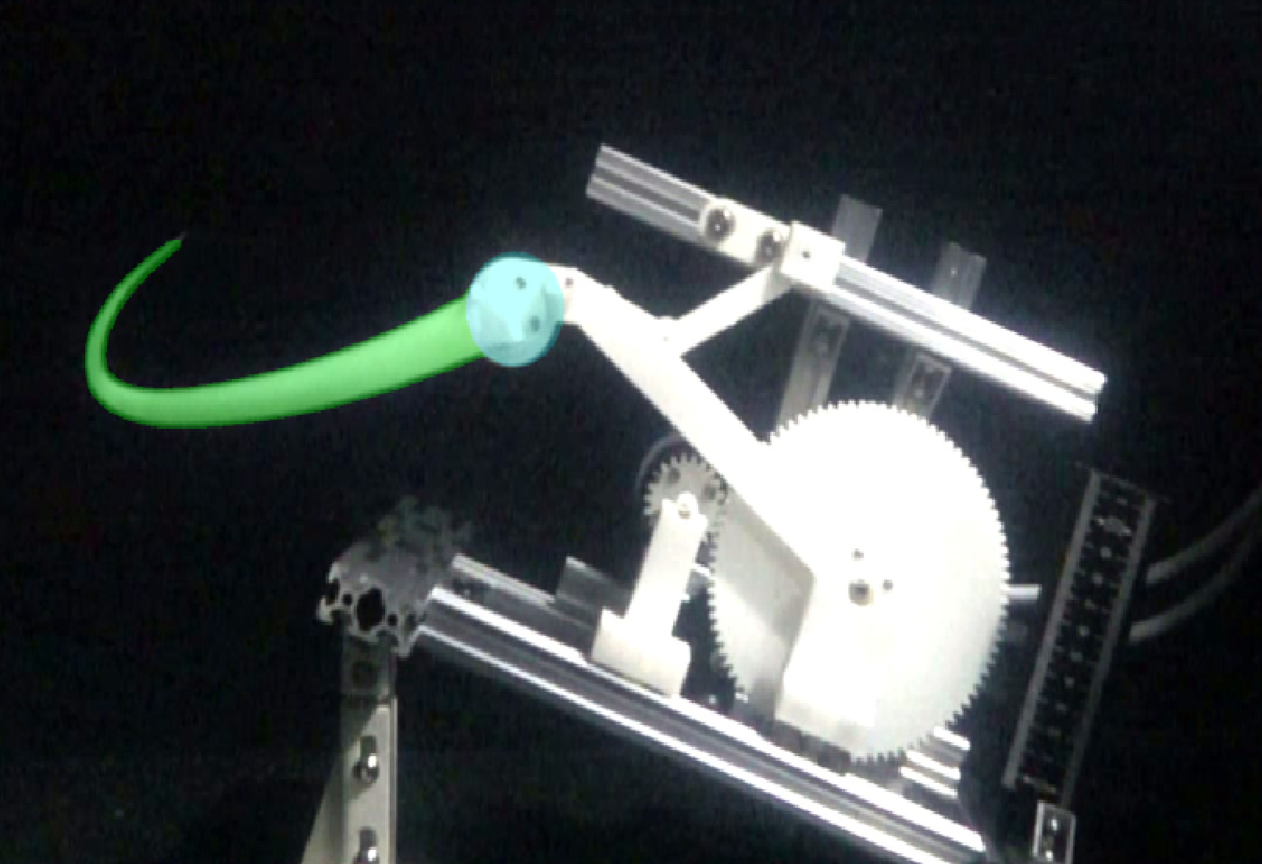}}%
  \hfill
  \subfloat[]{%
    \label{fig:Fig4b}
    \includegraphics[width=0.48\linewidth]{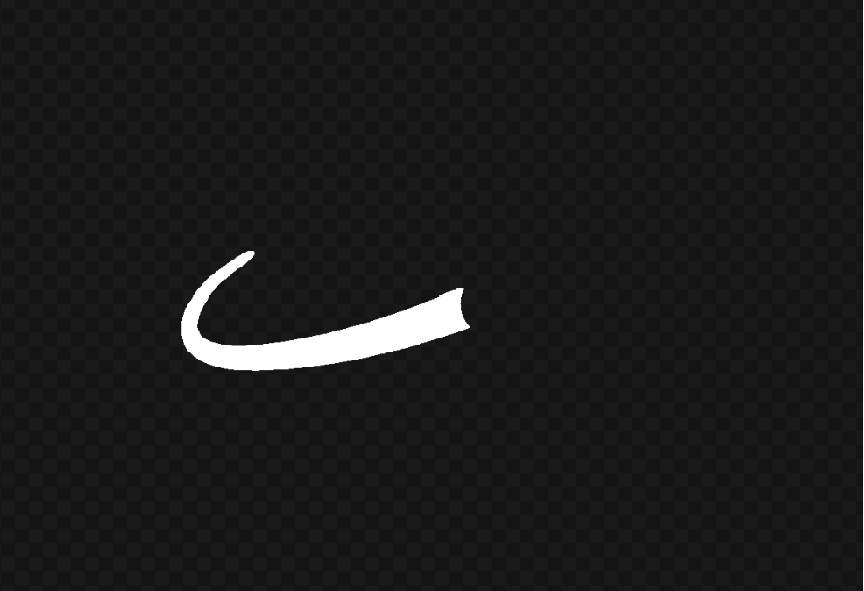}}%
  \vspace{-3pt}
  \centering
  \subfloat[]{%
    \label{fig:Fig4c}
    \includegraphics[width=0.48\linewidth]{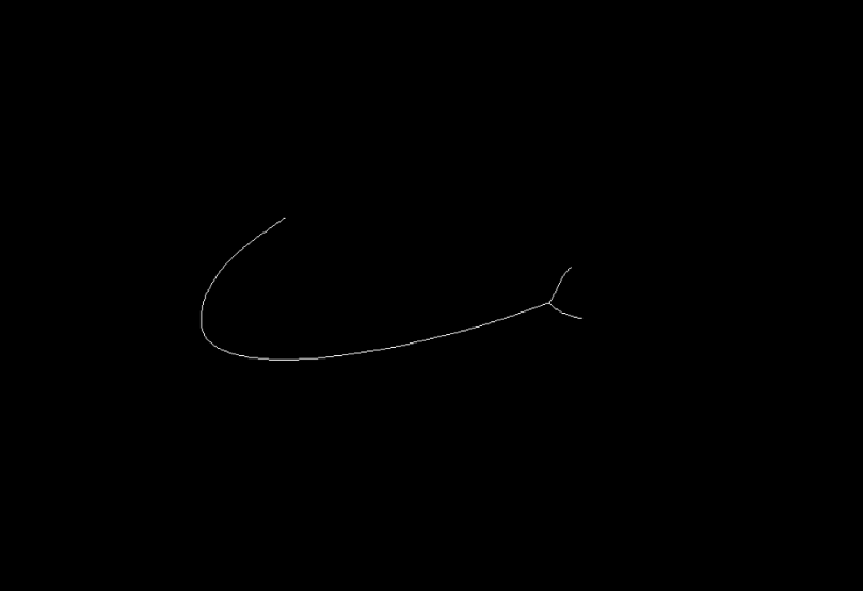}}%
  \hfill
  \subfloat[]{%
    \label{fig:Fig4d}
    \includegraphics[width=0.48\linewidth]{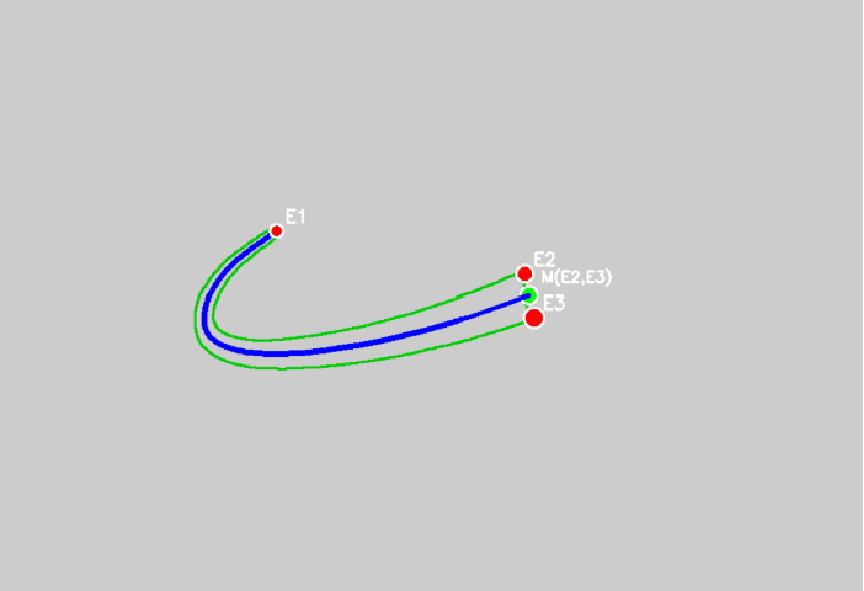}}%
  \caption{(a) Segmentation on original image. (b) Extracted mask. (c) Original midline with a branch. (d) Final midline.}
  \label{fig:Fig4}
\end{figure}
\textbf{Midline Extraction.}
To quantitatively analyze the deformation of the silicone octopus arm, we developed a custom image-processing pipeline. As shown in Figure.~\ref{fig:Fig4a}, each video frame was first processed with Meta’s Segment Anything Model~2 (SAM~2) \cite{ravi2024sam2} to segment the moving arm from the underwater background. A binary mask was then applied to remove the rigid supporting structure, yielding a clean extraction of the silicone arm (Figure.~\ref{fig:Fig4b}). The extracted region was skeletonized to obtain its medial axis (Figure.~\ref{fig:Fig4c}). Because skeletonization can occasionally introduce spurious branches, all branch points were detected and removed. The three terminal nodes of the remaining skeleton were subsequently identified, and the two closest endpoints $(E_2, E_3)$ were used to compute their midpoint $M$, which served as an auxiliary node to reconstruct the main, ordered centerline (Figure.~\ref{fig:Fig4d}).

\textbf{Curvature Computation.}
The reconstructed midline was smoothed and uniformly resampled to generate evenly spaced reference points along the arm. 
For each sampled point, the local curvature $\kappa$ was computed using a three-point geometric method: for a point $P_i$ and its two adjacent neighbors $P_{i-1}$ and $P_{i+1}$, a circle passing through these three points was fitted. 
This approach yields a continuous curvature profile along the arm, effectively describing the spatial propagation of bending.

\textbf{Bend-point Velocity Identification.}
The bending point was defined as the midline location with the minimum $x$-coordinate in each frame. Using the known time interval between consecutive frames, the displacement of this bending point was tracked to compute its velocity. This procedure provides a quantitative measure of the propagation speed of the curvature wave along the soft robotic arm during the reaching motion.

For a comprehensive and detailed account of the fabrication procedures, experimental setup, and physical modeling, please refer to \cite{zsy}.

\section{RESULTS}
\begin{figure*}[!b]
    \centering
    \includegraphics[width=\textwidth]{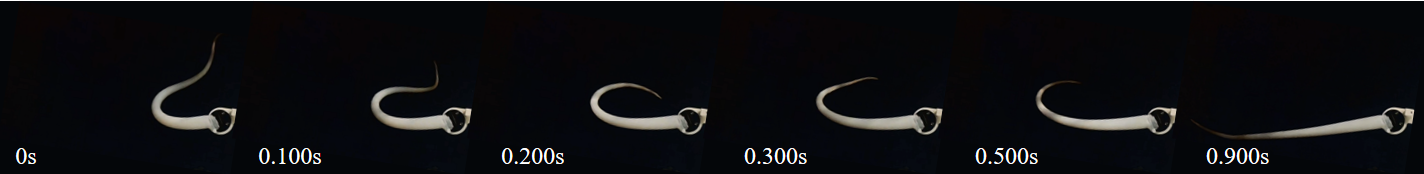}
    \caption{Ecoflex Gel 2 underwater performance at 150 rpm.}
    \label{fig:Fig5}
\end{figure*}
\begin{figure*}[!b]
    \centering
    \subfloat[Normalized velocity pattern for velocity and distance and aligned at the peaks.\cite{gutfreund1996organization}]{%
    \label{fig:Fig6a}
    \includegraphics[width=0.5\textwidth]{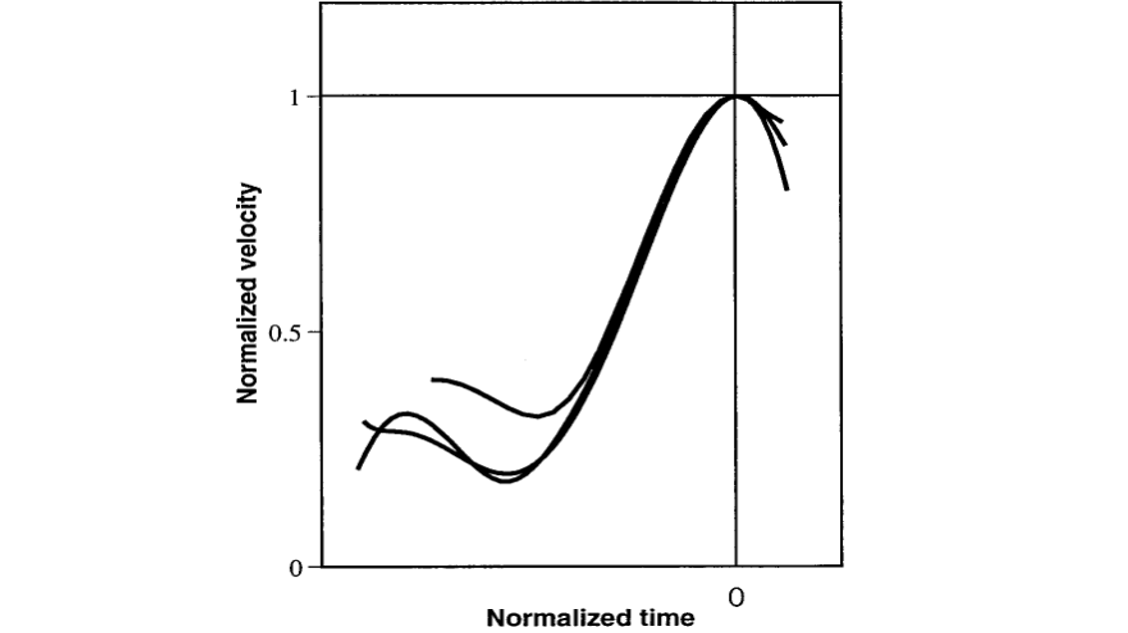}
}\hfill
    \subfloat[Normalized velocity pattern in the experiment.]{%
    \label{fig:Fig6b}
        \includegraphics[width=0.46\textwidth]{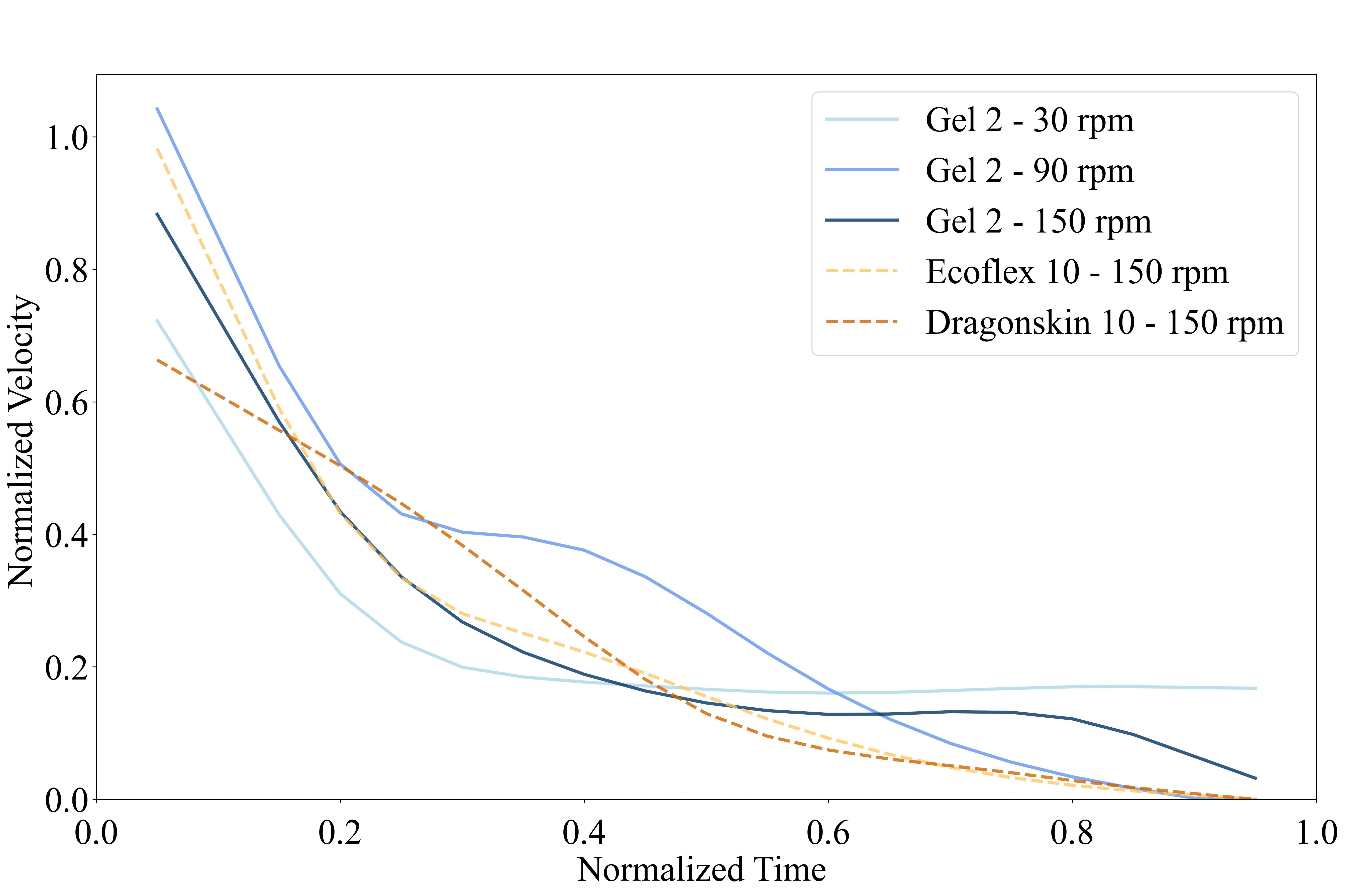}
    }\hfill
    \caption{Curvature and bend point velocity characteristics of bend propagation.}
    \label{fig:Fig6}
\end{figure*}

Due to the limited availability of quantitative data on the mechanical properties of live octopus arms—properties that are known to vary with species, age, and physiological condition—it was not feasible to target an exact stiffness match to biological tissue. Material selection was therefore informed by qualitative tactile comparison \cite{janardhana2025comprehensive}, among which Gel 2 silicone (the softest silicone we can find) exhibited the most distinct and similar bend propagation response. Based on this observation, Gel 2 was chosen as the reference material for subsequent variable-speed experiments to identify a driving velocity capable of generating clear bend propagation. After this optimal condition was established, the same fixed-speed tests were conducted on Ecoflex 10 and Dragonskin 10 for comparison.

\begin{figure}[!b]
    \centering
    \subfloat[Ecoflex Gel 2 at 30 rpm]{%
        \label{fig:Fig7a}
        \includegraphics[width=0.67\linewidth]{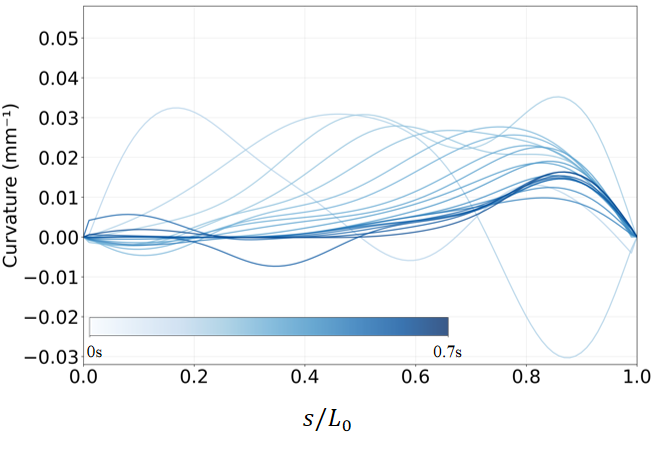}
    }
    \vspace{0pt} 
    \subfloat[Ecoflex Gel 2 at 90 rpm]{%
        \label{fig:Fig7b}
        \includegraphics[width=0.67\linewidth]{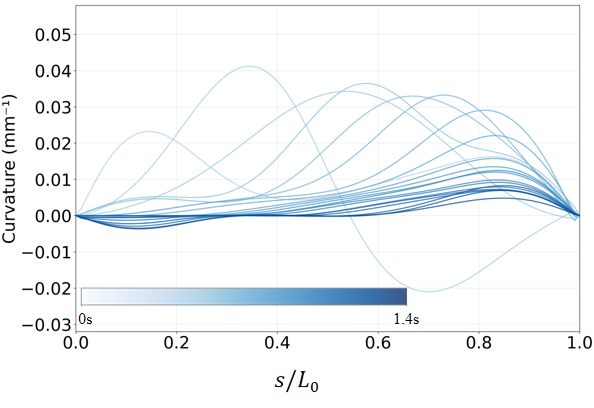}
    }
    \vspace{0pt} 
    \subfloat[Ecoflex Gel 2 at 150 rpm]{%
        \label{fig:Fig7c}
        \includegraphics[width=0.71\linewidth]{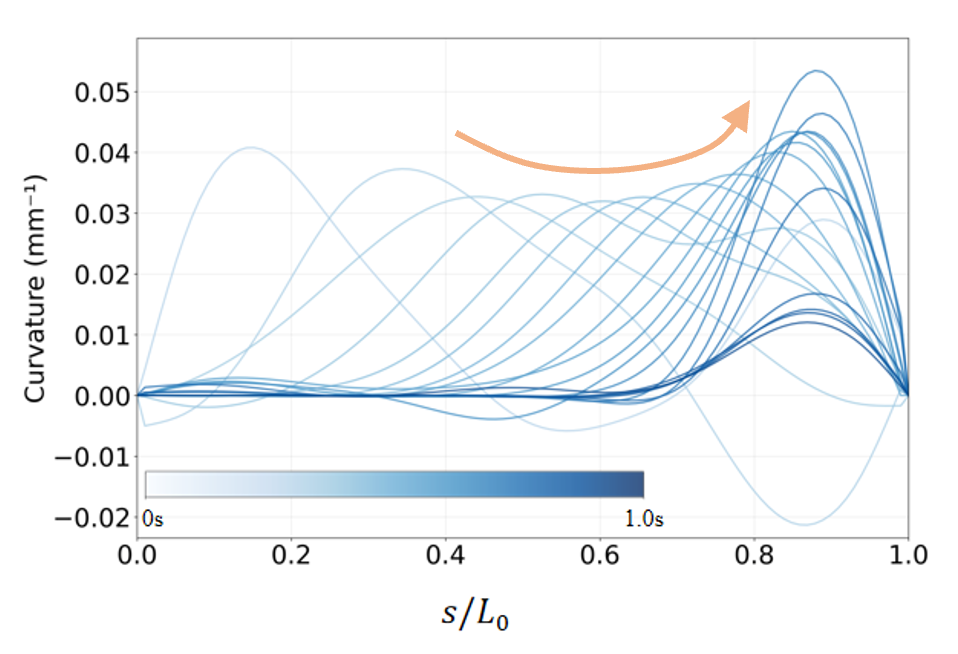}
    }
    \vspace{0pt}
    \subfloat[Ecoflex 10 at 150 rpm]{%
        \label{fig:Fig7d}
        \includegraphics[width=0.67\linewidth]{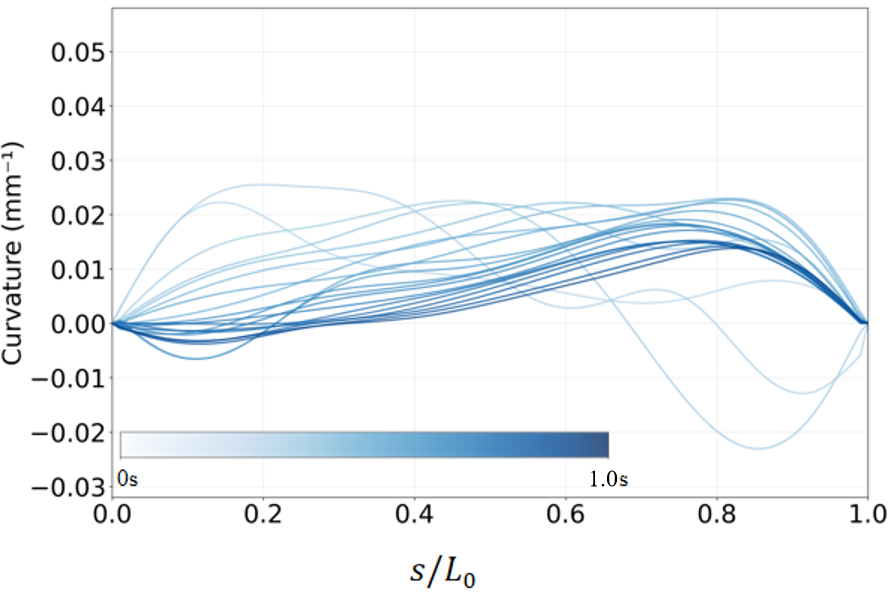}
    }
    \vspace{0pt}   
    \subfloat[Dragonskin 10 at 150 rpm]{%
        \label{fig:Fig7e}
        \includegraphics[width=0.67\linewidth]{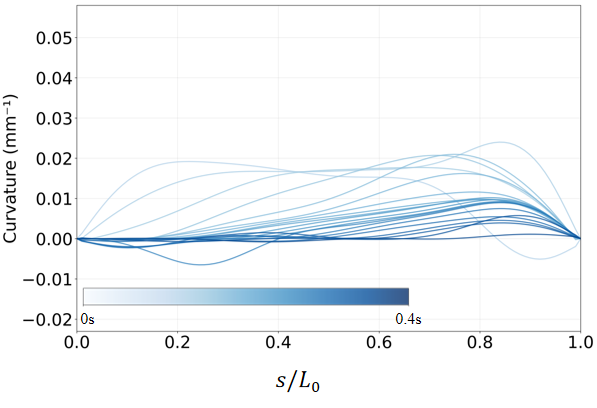}
    }
    \caption{Curvature characteristics of bend propagation under different driving speeds and materials.}
    \label{fig:Fig7}
\end{figure}

In the variable-speed tests, three driving motor velocities were evaluated: 30 rpm, 90 rpm, and 150 rpm. Motions exceeding 210 rpm were excluded because the arm exhibited out-of-plane deviations that disrupted the planar curvature propagation. For each condition, the recorded trajectories were processed to visualize the bend point velocity as a function of normalized time and velocity and the temporal evolution of curvature along the normalized arm length, as shown in Figure.~\ref{fig:Fig6b} and Figure.~\ref{fig:Fig7}.

For different driving speeds of the Gel 2 arm (Figure.~\ref{fig:Fig7a}–\ref{fig:Fig7c}), the curvature pattern at 150 rpm was most consistent with biological reaching (see Figure.~\ref{fig:Fig5}), showing a temporal evolution similar to that observed in octopus arm motion (Figure.~\ref{fig:intro_b}). The curvature first decreased, then increased, and finally decayed, resulting in the largest amplitude and most distinct propagation among the samples tested. In contrast, Ecoflex 10 and Dragonskin 10 produced smaller curvature magnitudes and less continuous propagation  (Figure.~\ref{fig:Fig7c}–\ref{fig:Fig7e}).

In both the variable-speed experiments (Gel 2 at 30, 60, and 150 rpm) and the variable-material experiments (Gel 2, Ecoflex 10, and Dragonskin 10 at a fixed speed of 150 rpm), the propagation velocity of the bend point along the x-direction exhibited a monotonically decreasing trend over time (Figure.~\ref{fig:Fig6b}). In all cases, the bend point started with a relatively high velocity, which rapidly decayed and gradually approached zero. 

\section{DISCUSSION}

\begin{figure*}[!t]
    \centering
    \subfloat[]{%
    \label{fig:Fig8a}
    \includegraphics[width=0.9\linewidth]{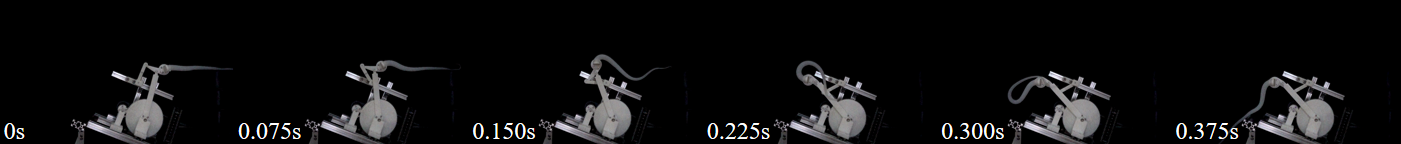}}%
\\
    \subfloat[]{%
    \label{fig:Fig8b}
    \includegraphics[width=0.9\linewidth]{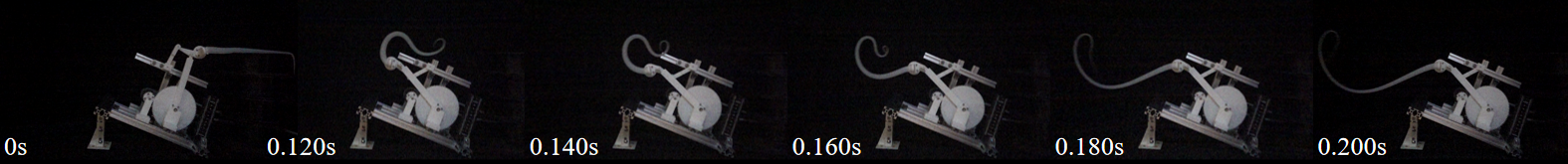}}%
\caption{Ecoflex Gel 2 performance in air at (a) 150 rpm (b) 330 rpm.}
    \label{fig:Fig8}
\end{figure*}

Material and speed strongly modulate the bend propagation. Among the tested conditions, Gel 2 generated the largest curvature amplitudes and the most continuous transmission; Ecoflex 10 showed intermediate behavior, while Dragonskin 10 attenuated fastest. Driving at 150 rpm provided an optimal balance between input energy and hydrodynamic loading, producing the most similar propagation to biological reaching. By contrast, lower speeds yielded insufficient momentum to sustain wave transmission, whereas excessive driving speeds introduced inertial and hydrodynamic instabilities that disrupted the planar mode of propagation.

Despite the clear spatial resemblance between the curvature evolution of Gel 2 at 150 rpm and biological reaching, the bend-point velocity shows a monotonically decreasing time course in both the variable-speed and variable-material experiments (see Figure.~\ref{fig:Fig6b}) instead of the bell-shaped profile typical of biological reaching as shown in Figure.~\ref{fig:Fig6a}. This finding supports the interpretation that the observed motion represents a passive hydrodynamic propagation rather than an actively controlled reaching behavior. Mechanistically, the monotonic decay accords with dissipative propagation of a passive wave, where viscous drag from the fluid continuously drain energy: the initial actuation originates at the base and attenuates as it travels, leading to a reduction of kinetic energy and curvature amplitude.

Interestingly, the curvature patterns of Gel 2 at low driving speed (30 rpm) exhibited temporal profiles similar to those of Ecoflex 10 and Dragonskin 10 driven at 150 rpm. This correspondence suggests that curvature propagation depends not only on material modulus but also on the relative balance between actuation intensity and inertial loading. The softer Gel 2 at low driving speed and the stiffer Ecoflex 10 and Dragonskin 10 at moderate speed appear to reach comparable dynamic regimes where elastic recovery dominates over inertia-driven momentum transfer. Under these conditions, the bending wave travels through passive redistribution of kinetic and elastic energy along the arm, leading to similar curvature attenuation profiles across materials. This behavior aligns with the whip-wave dynamics described by McMillen and Goriely \cite{mcmillen2003whip}, where energy imparted at the base propagates through a continuous medium via inertial coupling and geometric nonlinearity.

Furthermore, at higher speeds (210 rpm), the silicone arm occasionally exhibited non-planar motion characterized by twisting and asymmetric trajectories. We hypothesize that this phenomenon arises from strong fluid–structure coupling, where the interaction between the rapidly deforming arm and the surrounding flow generates three-dimensional hydrodynamic instabilities. Similar out-of-plane deformations have been reported in previous fluid–structure interaction studies of flexible bodies in oscillatory or traveling-wave motion \cite{wang2020review,ma2022flexible}. In these works, the coupling between structural elasticity, inertial loading, and unsteady vortex shedding can drive spontaneous transitions from planar to helical or chaotic motion once the driving velocity exceeds a critical threshold. In our system, this transition likely reflects the onset of such coupling-induced instability, suggesting that fluid feedback becomes dynamically significant beyond the regime of quasi-planar propagation.

These findings indicate that the bend propagation observed in our experiments can be parsimoniously explained by a passive hydrodynamic mechanism. Once a local bend is initiated near the base, distributed drag from the surrounding water effectively suppresses the unactuated portions of the arm and advects the bend distally along a nearly planar path. The absence of propagation in air under identical conditions (Gel 2 driven at 150 rpm), as shown in Figure.~\ref{fig:Fig8a}, further supports this interpretation. In addition, experiments in the attached video conducted in air with driving speeds at 210, 270, and 330 rpm (see Figure.~\ref{fig:Fig8b}) revealed bend propagation patterns that differed substantially from those observed in biological reaching, accompanied by a marked increase in bend-point velocity  \cite{mcmillen2003whip}. This suggests that, in the absence of fluid-mediated damping and added-mass effects, the motion is dominated by direct mechanical inertia rather than hydrodynamic coupling. Therefore, we speculate that the surrounding fluid plays an important role in shaping curvature propagation, consistent with previous biological findings describing passive wave transmission in “non-activated” octopus arm segments \cite{yekutieli2005dynamic}.

\section{CONCLUSIONS}

This study examined whip-like motions in a fluid environment and found that, while a Gel 2 arm driven at 150 rpm produces curvature patterns similar to octopus reaching, its bend-point velocity decays monotonically—lacking the bell-shaped profile characteristic of biological reaching. The absence of bend propagation in air further supports that the observed motion is governed primarily by buoyancy, passive hydrodynamics, and fluid–structure interactions. These results offer a complementary perspective: reaching is not simply a whipping behavior.

From a robotics perspective, this mechanism is promising for soft and underwater control. Unlike conventional multi-DOF arms requiring coordinated, control-intensive actuation, the single-motor design enables faster, more compliant, and energy-efficient directional extension. This approach is especially suited to low-DOF underwater grasping, where multiple motors pose spatial and waterproofing challenges. Integrating suckers or adhesives onto the passive-propagation arm could yield compact, efficient manipulation systems for confined underwater operations.

The proposed methodology—combining controlled actuation, underwater trials, and quantitative kinematic analysis—provides a simple and reproducible platform for studying reaching-like motion. It may support systematic investigations into the effects of fluid viscosity, arm geometry, and input shaping. Future work will explore drive profiles and stiffness distributions to approximate bell-shaped velocity profiles without compromising propagation robustness, and extend analysis to 3D kinematics at higher speeds.

\bibliographystyle{IEEEtran}
\bibliography{Mybib}

\end{document}